%% file: main.tex
\documentclass[nohyperref]{article}

\input{math_commands.sty}

\usepackage[accepted]{icml2023}

\definecolor{red}{RGB}{207,78,83}
\definecolor{blue}{RGB}{87,148,160}
\icmltitlerunning{Unsupervised Learning on a \textbf{DIET}}

\begin{document}

\twocolumn[
\icmltitle{Unsupervised Learning on a DIET: Datum IndEx as Target\\Free of Self-Supervision, Reconstruction, Projector Head}



\icmlsetsymbol{equal}{*}

\begin{icmlauthorlist}
\icmlauthor{Randall Balestriero}{a}
\end{icmlauthorlist}

\icmlaffiliation{a}{Meta AI, FAIR, NYC, USA}

\icmlcorrespondingauthor{}{rbalestriero@meta.com}

\icmlkeywords{Machine Learning, ICML}

\vskip 0.3in
]



\printAffiliationsAndNotice{}



\author{Antiquus S.~Hippocampus, Natalia Cerebro \& Amelie P. Amygdale \thanks{ Use footnote for providing further information
about author (webpage, alternative address)---\emph{not} for acknowledging
funding agencies.  Funding acknowledgements go at the end of the paper.} \\
Department of Computer Science\\
Cranberry-Lemon University\\
Pittsburgh, PA 15213, USA \\
\texttt{\{hippo,brain,jen\}@cs.cranberry-lemon.edu} \\
\And
Ji Q. Ren \& Yevgeny LeNet \\
Department of Computational Neuroscience \\
University of the Witwatersrand \\
Joburg, South Africa \\
\texttt{\{robot,net\}@wits.ac.za} \\
\AND
Coauthor \\
Affiliation \\
Address \\
\texttt{email}
}

%

\newcommand{\fix}{\marginpar{FIX}}
\newcommand{\new}{\marginpar{NEW}}

\begin{abstract}
\input{content/abstract}
\end{abstract}

\input{content/introduction}

\input{content/background}


\section{Unsupervised Learning on a DIET}

We first present in \cref{sec:DIET} the DIET enabling competitive yet simple unsupervised learning; in particular, \cref{sec:SOTA,sec:ablation} will demonstrate how that DIET competes and sometimes outperforms SSL on a variety of small to medium scale datasets while working out-of-the-box across architectures. \cref{sec:scaling} will then demonstrate how to scale DIET to large datasets such as Imagenet and INaturalist.

\subsection{The DIET: Datum IndEx as Target}
\label{sec:DIET}

The goal of this section is to introduce the proposed DIET, focusing on its simplicity and ease of implementation. Empirical validation of the method is deferred to the subsequent \cref{sec:SOTA,sec:ablation,sec:scaling} as summarized as the end of this section.

{\bf Algorithm:}~As well indicated by its name, Datum IndEx as Target (DIET) proposes to perform unsupervised learning by employing the datum index as its target class. That is, given a dataset of $N$ samples $\{\vx_1,\dots,\vx_N\}$, define the class of sample $\vx_n$ with $n \in \{1,\dots,N\}$ to be $n$ itself, leading to the DIET's loss $\mathcal{L}_{\rm\small DIET}$ for a datum to be
\begin{align}
    \mathcal{L}_{\rm\small DIET}(\vx_n)={\rm ClassificationLoss}({\color{red}\mW}{\color{blue}f_{\vtheta}}(\vx_n),n),\label{eq:DIET}
\end{align}
given a sample $\vx_n \in \mathbb{R}^{D}$, a DN ${\color{blue}f_{\vtheta}:\mathbb{R}^{D} \mapsto \mathbb{R}^{K}}$, DIET's linear classifier ${\color{red}\mW \in \mathbb{R}^{K \times N}}$, and one's preferred classification loss. Unless one expects to sample some data more than others, there is no reason to add a bias vector to DIET's linear classifier. As such, DIET performs unsupervised learning through a supervised scheme meaning that any progress made within the supervised learning realm can be ported as-is to DIET. Throughout our study, we will be employing the cross-entropy loss, denoted as \texttt{X-Ent}. We summarize DIET in \cref{fig:DIET} and propose its pseudo-code in \cref{algo:DIET} as well as the code to obtain a data loader providing the user with the indices ($n$) in \cref{algo:dataset}.

\begin{algorithm}[t!]
\begin{lstlisting}[language=Python,escapechar=\%,numbers=none]
# take any preferred DN e.g. resnet50
# see %\lsComment{\cref{algo:architectures}}% for other examples
f = torchvision.models.resnet50() # %\lsComment{\color{blue}$f_{\vtheta}$}% in %\lsComment{\cref{eq:DIET}}%

# f comes with a classifier, let's remove it
K = f.fc.in_features # = %\lsComment{\color{blue}$f_{\vtheta}$}%'s output dim.
f.fc = nn.Identity() # remove it so f=%\lsComment{\color{blue}$f_{\vtheta}$}% in %\lsComment{\cref{eq:DIET}}%

# define DIET's linear classifier and X-Ent
W = nn.Linear(K, N, bias=False) # %\lsComment{\color{red}$\mW$ in \cref{eq:DIET}}%
diet_loss = nn.CrossEntropyLoss(label_smoothing=0.8)

# start DIET training %\lsComment{(Fig.~\ref{fig:DIET})}% y is optional
for x, n in train_loader: # see %\lsComment{\cref{algo:dataset}}% for loader
    preds = f(t(x)) # t applies the DA %\lsComment{($\mathcal{T}$ in Fig.~\ref{fig:DIET})}%
    loss = diet_loss(W(preds), n) # %\lsComment{\cref{eq:DIET}}%
    # proceed with backprop/optimizer/scheduler
\end{lstlisting}
\caption{\small \ul{DIET's algorithm}, minimal code refactoring is required to employ the DIET given any already built deep learning pipeline, to obtain a dataset that provides the indices ($n$), see \cref{algo:dataset}, (\texttt{nn} stands for \texttt{torch.nn}, Pytorch used for illustration).}
\label{algo:DIET}
\end{algorithm}
\begin{algorithm}[t!]
\begin{lstlisting}[language=Python,escapechar=\%,numbers=none]
from torch.utils.data import Dataset, DataLoader

class DatasetWithIndices(Dataset):
    def __init__(self, dataset):
        self.dataset = dataset
    def __getitem__(self, n):
        # if true class label is given, disregard it
        x, _ = self.dataset[n]
        return x, n
    def __len__(self):
        return len(self.dataset)

# example of train_loader with CIFAR100
CIFAR100 = torchvision.datasets.CIFAR100(root)
CIFAR100_w_ind = DatasetWithIndices(CIFAR100)
train_loader = DataLoader(C100_w_ind)
\end{lstlisting}
\caption{\small \ul{Custom loader} (\texttt{train\_loader} in \cref{algo:DIET}) to obtain the indices ($n$) in addition to the inputs $\vx_n$ and (optionally) the labels $y_n$ (Pytorch used for illustration).}
\label{algo:dataset}
\end{algorithm}

\begin{figure*}[t!]
    \centering
    \includegraphics[width=0.32\linewidth]{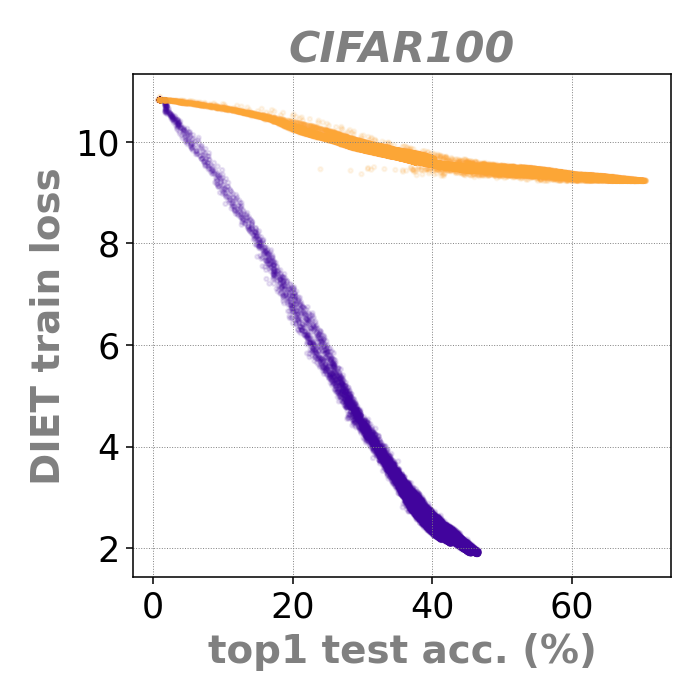}
    \includegraphics[width=0.32\linewidth]{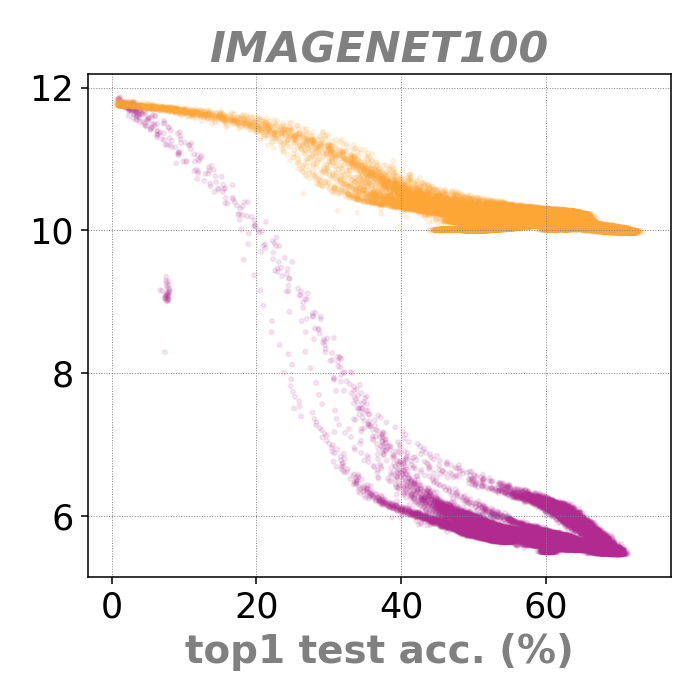}
    \includegraphics[width=0.32\linewidth]{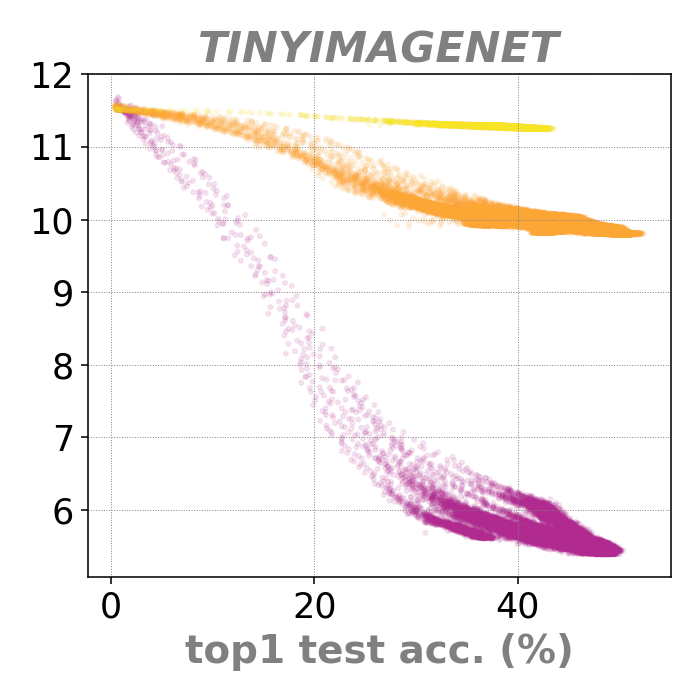}\\[-1em]
    \caption{Depiction of the DIET's training loss ({\bf y-axis}) against the online test linear probe accuracy ({\bf x-axis}) for all the models and hyper-parameters corresponding to the experiments of \cref{tab:C100} for CIFAR100 ({\bf left column}), and \cref{tab:tiny} for IN100 and TinyImagenet ({\bf middle and left columns}). We colorize ({\bf yellow to blue}) the points based on the strength of the label smoothing parameter (recall that it plays a role in DIET's convergence speed from \cref{sec:ablation}). We clearly identify that for a given label smoothing parameter, there exists a strong relationship between \textbf{DIET}'s training loss and the test accuracy showing that model selectin can be performed this way. Therefore,  even without labels, \textbf{DIET}'s loss can be used as a quantitative quality assessment measure of one's model. The shift observed for different values of label smoothing can be accounted for to re-calibrate all the experiment if needed using the known relationship between that increasing the label smoothing parameter decreases the X-Ent loss, everything else equal.}
    \label{fig:accus}
\end{figure*}

{\bf Benefits:}~We ought to highlight three key benefits of DIET's \cref{eq:DIET}. First, the amount of code refactoring is minimal (recall \cref{algo:DIET}): there is no change to be done into the data loading pipelines (recall \cref{algo:dataset}) as opposed to SSL which requires positive pairs, no need to specify teacher-student architectures, and no need to design a projector/predictor DN. Second, DIET's implementation is not architecture specific as we will validate on Resne(x)ts, ConvNe(x)ts, Vision Transformers and their variants. Third, DIET does not introduce any additional hyper-parameters in addition to the ones already present in one's favorite {\rm ClassificationLoss} used in \cref{eq:DIET}, all while providing a training loss that is informative of test time performances, as we depict in \cref{fig:accus}. 

{\bf Relation to previous methods:}~Despite DIET's simplicity, we could not find an existing method that considered it perhaps due to the common --but erroneous-- belief that dealing with hundreds of thousands of classes (N in \cref{fig:DIET}, the training set size) would not produce successful training. As such, the closest method to ours is Exemplar CNN \cite{alexey2015discriminative} which extracts a few patches from a given image dataset, and treat each of them as their own class; this way the number of classes is the number of extracted patches, which is made independent from N. Performances are however far below SSL methods. A more recent method, Instance Discrimination \cite{wu2018unsupervised} extends Exemplar CNN by introducing inter-sample discrimination. However, they do so using a non-parametric softmax {\em i.e.} by defining a learnable bank of centroids to cluster training samples; for successful training those centroids must be regularized to prevent representation collapse. As we will compare in \cref{tab:C100}, DIET outperforms Instance Discrimination while being simpler. Lastly, methods such as Noise as Targets \cite{bojanowski2017unsupervised} and DeepCluster \cite{caron2018deep} are quite far from DIET as (i) they perform clustering and use the datum's cluster as its class {\em i.e.} greatly reducing the dependency on N, and (ii) they perform such clustering in the output space of the model {\color{blue} $f_{\vtheta}$} being learned which brings multiple collapsed solutions requiring those methods to employ complicated mechanisms to ensure training to learn non-trivial representations. 

{\bf Empirical validation roadmap:}~To support the different claims we have made above, we propose to scatter our empirical validation in a few subsequent sections. First, we will explore small and medium scale datasets in \cref{sec:SOTA} that include the eponymous CIFAR100 but also other datasets such as Food101 which have been challenging for SSL. In fact, whenever the number of training samples is small, most SSL methods favor Imagenet pre-training and few-shot transfer learning. In \cref{sec:SOTA} we will also consider TinyImagenet and Imagenet-100. After having validated the ability of DIET to match and sometimes outperform SSL methods, we will spend \cref{sec:ablation} to probe the few hyper-parameters that govern DIET, in our case the label smoothing of the \texttt{X-Ent} loss, and the training time. We will see that without label smoothing, DIET is often as slow as SSL methods to converge, and sometimes slower --but that high values of label smoothing greatly speed up convergence. Lastly, we dedicate \cref{sec:scaling} to scaling up DIET to large dataset such as Imagenet and INaturalist. In fact, recall from \cref{fig:DIET} that DIET's N-output classifier becomes a memory bottleneck when $N$ is large in which case a slightly different treatment is required to employ DIET. We will see that even the most naive solutions {\em e.g.} subsampling of a large dataset enables to apply DIET as-is all while producing highly competitive performances.

\begin{figure}[t!]
\begin{framed}
\small
    \textbf{DIET's experimental setup:}
    \begin{itemize}[leftmargin=3.5mm,itemsep=-1pt,topsep=0.pt]
        \item \color{blue}Official Torchvision architectures \color{black} (\ul{no changes in init./arch.}), only swapping the classification layer with \color{red}DIET's one \color{black} (right of \cref{fig:DIET}), \ul{no projector DN}
        \item \ul{Same DA pipeline} ($\mathcal{T}$ in \cref{fig:DIET}) across datasets/architectures with \ul{batch size of 256} to fit on 1GPU
        \item \ul{AdamW optimizer with linear warmup (10 epochs) and cosine annealing} learning rate schedule, \ul{\texttt{X-Ent} loss} (right of \cref{fig:DIET}) with \emph{label smoothing of $0.8$} 
        \item \emph{Learning rate/weight-decay} of $0.001/0.05$ for non transformer architectures and $0.0002/0.01$ for transformers
    \end{itemize}
    \vspace{-0.2cm}
    \caption{In \ul{underlined} are the design choices directly ported from standard supervised learning (not cross-validated for DIET), in \emph{italic} are the design choices cross-validated for DIET but held constant across this study unless specified otherwise. Batch-size sensitivity analysis is reported in \cref{tab:BS_CV,fig:BS_CV} showing that performances do not vary when taking values from $32$ to $4096$. \texttt{X-Ent}'s label smoothing parameter plays a role into DIET's convergence speed, and is cross-validated in \cref{fig:epochs,tab:epochs}; we also report DA ablation in \cref{fig:DA,tab:DA}.}
    \label{fig:hparams}
    \end{framed}
    \vspace{-0.5cm}
\end{figure}

Throughout our subsequent empirical validation, we will religiously follow the experimental setup described in \cref{fig:hparams}, unless stated otherwise. Our goal in adopting the same setup across experiments is to highlight the stability of DIET to dataset and architectural changes; careful tuning of those design choices should naturally lead to greater performance if desired.

\subsection{The DIET Competes with the State-Of-The-Art}
\label{sec:SOTA}

We start the empirical validation of DIET on the eponymous CIFAR100 dataset; following that, we will consider other common medium scale datasets {\em e.g.} TinyImagenet, and in particular we will consider datasets such as Food101, Flowers102 for which current SSL does not provide working solutions and for which the common strategy consists in transfer learning. We will see in those cases that applying DIET as-is on each dataset is able to produce high-quality representations for a large set of DN architectures.

\begin{table}[t!]
    \centering
    \setlength{\tabcolsep}{0.32em}
    \renewcommand{\arraystretch}{0.6}
    \caption{{\bf CIFAR100/linear-probe/single-GPU}: with the settings of \cref{fig:hparams} and optionally longer training (LT) or different label-smoothing (ls) specified, notice the consistent progression of the performance through architectures which is not easily achieved in SSL. In particular, we recall that the alternative methods all employ (except MoCo) a projector network (recall \cref{sec:background}).
    Benchmarks taken from
    $\dagger:$\citet{ho2020contrastive};
    $\ddagger:$\citet{peng2022crafting};
    $\ast$:\citet{zhang2022dual};
    $\bullet$:\citet{pham2022pros};
    $\diamond$:\citet{da2022solo};
    $\star$:\citet{yeh2022decoupled};
    $\triangleleft$:\citet{ren2022simple};
    $\triangleright$:\citet{yang2022identity};
    $\Box$:\citet{huang2022self}.
    }
    \label{tab:C100}
    \begin{tabular}{@{}c@{}|@{}c@{}}
    \begin{tabularx}{0.5\linewidth}{>{\hsize=1.5\hsize}X | >{\hsize=.5\hsize}X}
        \multicolumn{2}{c}{\em\cellcolor{blue!45}Resnet18}\\
        MoCoV2& 53.28$^\ast$\\
        SimSiam& 53.66$^\bullet$\\
        SimCLR&53.79$^\dagger$\\ 
        SimMoCo& 54.11$^\ast$\\
        ReSSL&54.66$^\bullet$\\
        SimCLR+adv & 55.51$^\dagger$\\
        MoCo& 56.10$^\ddagger$\\
        SimCLR&56.30$^\star$\\
        MoCo+CC& 57.65$^\ddagger$\\
        SimCLR&57.81$^\triangleright$\\
        DINO& 58.12$^\bullet$\\
        SimCO& 58.35$^\ast$\\
        SimCLR+DCL & 58.50$^\dagger$\\
        SimCLR&60.30$^\ddagger$\\
        SimCLR&60.45$^\bullet$\\
        W-MSE& 61.33$^\diamond$\\
        SimCLR+CC&61.91$^\ddagger$\\
        BYOL& 62.01$^\bullet$\\
        MoCoV2& 62.34$^\bullet$\\
        \cellcolor{red!15}DIET& \cellcolor{red!15}62.93\\
        BYOL& 63.75$^\ddagger$\\
        \cellcolor{red!15}DIET (LT)& \cellcolor{red!15}63.77\\
        BYOL+CC& 64.62$^\ddagger$\\
        SimSiam& 64.79$^\ddagger$\\
        SwAV& 64.88$^\diamond$\\
        SimCLR&65.78$^\diamond$\\
        SimSiam+CC& 65.82$^\ddagger$\\ 
    \end{tabularx}
    &
    \begin{tabularx}{0.5\linewidth}{X | >{\hsize=.5\hsize}X}
        \multicolumn{2}{c}{\em\cellcolor{blue!45}Resnet50}\\
        SimCLR& 52.04$^\dagger$\\
        MoCoV2& 53.44$^\ast$\\
        SimMoCo& 54.64$^\ast$\\
        SimCLR+adv& 57.71$^\dagger$\\
        SimCO& 58.48$^\ast$\\
        SimCLR& 61.10$^\star$\\
        SimCLR+DCL& 62.20$^\star$\\
        \cellcolor{red!15}DIET&\cellcolor{red!15}68.96\\
        MoCoV3&69.00$^\triangleleft$\\
        \cellcolor{red!15}DIET (ls:0.95)&\cellcolor{red!15}69.56\\
        \cellcolor{red!15}DIET (LT)&\cellcolor{red!15}69.91\\
        \multicolumn{2}{c}{\em\cellcolor{blue!45}Resnet101}\\
        SimCLR& 52.28$^\dagger$\\
        SimCLR+adv& 59.02$^\dagger$\\
        MoCoV3 &68.50$^\triangleleft$\\
        \cellcolor{red!15}DIET&\cellcolor{red!15}70.29\\
        \cellcolor{red!15}DIET (ls:0.95)&\cellcolor{red!15}71.09\\
        \cellcolor{red!15}DIET (LT)&\cellcolor{red!15}71.39\\
        \multicolumn{2}{c}{\em\cellcolor{blue!45}AlexNet}\\
        SplitBrain &39.00$^\Box$\\
        InstDisc &39.40$^\Box$\\
        DeepCluster& 41.90$^\Box$\\
        AND &47.90$^\Box$\\
        \cellcolor{red!15}DIET&\cellcolor{red!15}48.25\\
        SeLa&57.40$^\Box$\\
    \end{tabularx}
    \end{tabular}
    \vspace{-0.5cm}
\end{table}

\begin{figure*}[t!]
    \centering
    \includegraphics[width=\linewidth]{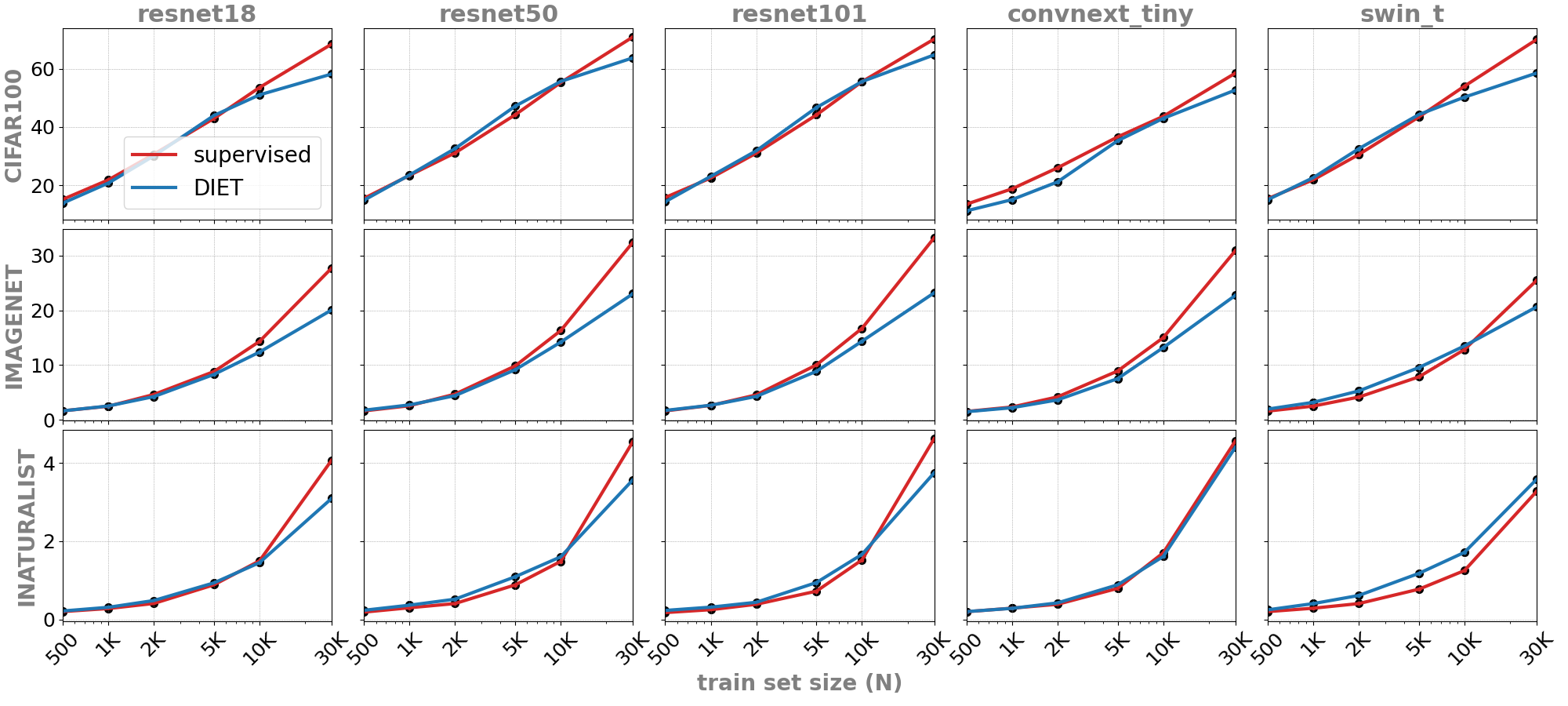}
    \caption{\small Depiction of DIET's performances ({\color{red} red}) against the supervised ({\color{blue} blue}) using a controlled training set size ({\bf x-axis}), subsampled from the original training set and identical between methods; evaluation is performed over the original full evaluation set and conducted on various dataset ({\bf rows}) and architectures ({\bf columns}). We see that for small training set size, DIET is able to match and sometimes even outperform the supervised benchmark. See \cref{fig:ablation} for additional datasets, in the Appendix.}
    \vspace{-0.3cm}
    \label{fig:ablation_short}
\end{figure*}

\begin{table}[t!]
    \setlength{\tabcolsep}{0.32em}
    \newcolumntype{P}[1]{>{\centering\arraybackslash}p{#1}}
    \centering
    \caption{{\bf TinyImagenet+IN100/linear-probe/single-GPU}: with the settings of \cref{fig:hparams}, as per \cref{tab:C100}, notice the consistent progression of the performance through architectures. We observe that DIET comes on par (higher-end for TinyImagenet and lower-end for IN100) with SSL methods.
    Benchmarks are taken from $1$:\cite{dubois2022improving}, $2:$\cite{ozsoy2022self}}
    \label{tab:tiny}
    \begin{tabular}{@{}c@{}|@{}c@{}}
    \multicolumn{2}{c}{\bf\ul{TinyImagenet}}\\
        \begin{tabularx}{0.5\linewidth}{X | X}\hline
         \multicolumn{2}{c}{\em\cellcolor{blue!45}Resnet18}\\
         SimSiam  & 44.54 $^\ddagger$\\
         \cellcolor{red!15}DIET     & \cellcolor{red!15}45.07\\
         SimCLR   & 46.21$^\ddagger$ \\
         BYOL     & 47.23$^\ddagger$ \\
         MoCo     & 47.98 $^\ddagger$\\
         SimClR   & 48.70 $^1$\\
         DINO     & 49.20 $^1$\\
        \end{tabularx}&
        \begin{tabularx}{0.5\linewidth}{X | X}\hline
         \multicolumn{2}{c}{\em\cellcolor{blue!45}Resnet50}\\
            SimCLr &48.12 $^2$\\
            SimSiam &46.76 $^2$\\
            Spectral &49.86 $^2$\\
            \cellcolor{red!15}DIET &\cellcolor{red!15}51.66\\
            CorInfoMax &54.86 $^2$\\
        \end{tabularx}\\ \hline
        \multicolumn{2}{c}{\cellcolor{red!15}DIET (other archs.)}\\
        \begin{tabularx}{0.5\linewidth}{>{\hsize=1.5\hsize}X | >{\hsize=.5\hsize}X}
        \em\cellcolor{blue!45} resnet34 & \cellcolor{red!15} 47.04 \\
        \em\cellcolor{blue!45} resnet101 & \cellcolor{red!15} 51.86 \\
        \em\cellcolor{blue!45} wide\_resnet50\_2 & \cellcolor{red!15} 50.03 \\
        \em\cellcolor{blue!45} resnext50\_32x4d & \cellcolor{red!15} 52.45 \\
        \em\cellcolor{blue!45} densenet121 & \cellcolor{red!15} 49.38 \\
        \end{tabularx}&
        \begin{tabularx}{0.5\linewidth}{>{\hsize=1.5\hsize}X | >{\hsize=.5\hsize}X}
            \em\cellcolor{blue!45} convnext\_tiny & \cellcolor{red!15} 50.88 \\
            \em\cellcolor{blue!45} convnext\_small & \cellcolor{red!15} 50.05 \\
            \em\cellcolor{blue!45} MLPMixer & \cellcolor{red!15} 39.32 \\
            \em\cellcolor{blue!45} swin\_t & \cellcolor{red!15} 50.80 \\
            \em\cellcolor{blue!45} vit\_b\_16 & \cellcolor{red!15} 48.38 \\
        \end{tabularx}\\
    \multicolumn{2}{c}{\bf\ul{Imagenet-100 (IN100)}}\\
    \begin{tabularx}{0.5\linewidth}{X | X}\hline
            \multicolumn{2}{c}{\em\cellcolor{blue!45}Resnet18}\\
            SimMoCo &58.20$^\ast$\\
            MocoV2  &60.52$^\ast$\\
            SimCo   &61.28 $^\ast$\\
         \cellcolor{red!15}DIET     & \cellcolor{red!15}64.31\\
            W-MSE2  &69.06 $^2$\\
            ReSSL   &74.02$^\bullet$\\
            DINO    &74.16$^\bullet$\\
            MoCoV2  &76.48$^\bullet$\\ 
            BYOL    &76.60$^\bullet$\\
            SimCLR  &77.04$^2$\\
            SimCLR  &78.72$^2$\\
            MocoV2  &79.28$^2$\\
            VICReg  &79.40$^2$\\
            Barlow  &80.38$^2$\\
        \end{tabularx}&
        \begin{tabularx}{0.5\linewidth}{>{\hsize=1.4\hsize}X | >{\hsize=.6\hsize}X}\hline
            \multicolumn{2}{c}{\em\cellcolor{blue!45}Resnet50}\\
         \cellcolor{red!15}DIET     & \cellcolor{red!15}73.50\\
            MoCo+Hyper.     &75.60 $^\star$\\
            MoCo+DCL        &76.80 $^\star$\\
            MoCoV2 + Hyper. &77.70 $^\star$\\
            BYOL            &78.76 $^2$\\
            MoCoV2 + DCL    &80.50 $^\star$\\
            SimCLR          &80.70 $^\star$\\
            SimSiam         &81.60$^2$\\
            SimCLR + DCL    &83.10 $^\star$\\
        \end{tabularx}\\
        \multicolumn{2}{c}{\cellcolor{red!15}DIET (other archs.)}\\
        \begin{tabularx}{0.5\linewidth}{>{\hsize=1.5\hsize}X | >{\hsize=.5\hsize}X}
        \em\cellcolor{blue!45} wide\_resnet50\_2 & \cellcolor{red!15} 71.92 \\
        \em\cellcolor{blue!45} resnext50\_32x4d & \cellcolor{red!15} 73.07 \\
        \em\cellcolor{blue!45} densenet121 & \cellcolor{red!15} 67.46 \\
            \em\cellcolor{blue!45} convnext\_tiny & \cellcolor{red!15} 69.77\\
        \end{tabularx}&
        \begin{tabularx}{0.5\linewidth}{>{\hsize=1.5\hsize}X | >{\hsize=.5\hsize}X}
            \em\cellcolor{blue!45} convnext\_small & \cellcolor{red!15} 71.06 \\
            \em\cellcolor{blue!45} MLPMixer & \cellcolor{red!15} 56.46 \\
            \em\cellcolor{blue!45} swin\_t & \cellcolor{red!15} 67.02 \\
            \em\cellcolor{blue!45} vit\_b\_16 & \cellcolor{red!15} 62.63 \\
        \end{tabularx}\\
    \end{tabular}
    \vspace{-0.5cm}
\end{table}

{\bf CIFAR100:}~Let's first consider CIFAR-100 \cite{krizhevsky2009learning} with a few variations of Resnet \cite{he2016deep} and AlexNet \cite{krizhevsky2014one}. To accomodate with the $32 \times 32$ resolution, we follow the standard procedure to slightly modify the ResNet architecture: the first convolution layer sees its kernel size go from 7$\times$7 to 3 $\times$ 3 and its stride reduce from 2 to 1; the max pooling layer following it is removed (details in \cref{algo:architectures}). On Alexnet, a few non-SSL baselines are available, and we thus compare with SplitBrain \cite{zhang2017split}, DeepCluster \cite{caron2018deep}, InstDisc \cite{wu2018unsupervised} (closest to ours, see \cref{sec:DIET}), AND \cite{huang2019unsupervised}, SeLa \cite{asano2019self}, ReSSL \cite{zheng2021ressl}. The models are trained and linear evaluation is employed to judge the quality of the learned model to solve the original classification task; results are reported in \cref{tab:C100}. We observe that DIET is able to match and often slightly exceed current SSL methods. In particular, even though CIFAR100 is a relatively small dataset, increasing the DN capacity {\em i.e.} going from Resnet18 to Resnet101 does not exhibit any overfitting.

{\bf TinyImagenet and IN-100:}~We continue our empirical validation by now considering the more challenging Imagenet100 (IN100) \cite{tian2020contrastive} dataset which consists of 100 classes of the full Imagenet-1k dataset, the list of classes can be found online\footnote{\url{https://github.com/HobbitLong/CMC/blob/master/imagenet100.txt}} and the TinyImagenet \cite{le2015tiny} dataset. Thanks to the higher resolution images present in those datasets, $224 \times 224$ and $64 \times 64$ respectively, we broaden the range of architecture we consider to include the Resnet variants of the previous section, SwinTransformer \cite{liu2021swin}, VisionTransform \cite{dosovitskiy2020image}, Densenet \cite{huang2017densely}, ConvNext \cite{liu2022convnet}, WideResnet \cite{zagoruyko2016wide}, ResNext \cite{xie2017aggregated}, and the MLPMixer \cite{tolstikhin2021mlp}. We report DIET and benchmark results in \cref{tab:tiny} where we now see that while on TinyImagenet, DIET is able to again strongly match SSL methods, DIET fall on the lower end of the spectrum on Imagenet100. Again we recall that as opposed to competing methods, DIET does not employ any projector DN.

\begin{table*}[t!]
    \centering
    \renewcommand{\arraystretch}{0.8}
    \setlength{\tabcolsep}{0.2em}
    \caption{Depiction of in-distribution and out-distribution (transfer) performances of DIET on a variety of datasets. They key observation is that even on small datasets such as Cars, Aircraft, DIET is able without any pretraining to come on-par with SSL performances obtained through Imagenet-1k pretraining. In addition, for medium scale dataset such as Food101, we see how DIET is able to significantly outperform SSL methods in both cases where we use DIET without pre-training, or when DIET is used to pretrain the architecture on another dataset. We also report performances for a ViT based architecture (SwinTiny) to demonstrate the ability of DIET to handle different models out-of-the-box, as per \cref{fig:hparams}, we only adapted the learning rate and weight decay parameters. Every DIET run in this table is done on 1GPU and using a batch-size of 128. $\dagger$:\cite{yang2022identity}, +:\cite{ericsson2021well}}
    \begin{tabular}{l|l|l|l|l|c|c|c|c|c|c}
       \multirow{3}{*}{Arch.} &\multirow{3}{*}{Pretrained} & \multirow{3}{*}{Frozen} & Method/Dataset&  {\em Aircraft}   & {\em DTD    }   & {\em Pets}      & {\em Flower}    & {\em CUB-200} & {\em Food101} &{\em StanfordCars}\\ 
       &&&N=&6667&1880&2940&1020&11788&68175&6509\\
       &&&C=&100&47&37&102&200&101&196\\\hline\hline
        \multirow{4}{*}{\em Resnet18}&\multirow{3}{*}{IN100$^\dagger$}&Yes&SimCLR        & 24.19       &   54.35   &   46.46   & 75.00     &16.73 &-&-\\
        &&&SimCLR + CLAE & 25.87       &   52.12   &   43.55   & 76.82     &17.58 &-&-\\
        &&&SimCLR + IDAA & 26.02       &   54.97   &   46.76   & 77.99     &18.15 &-&-\\\cline{2-11}
        &None&No&DIET & \cellcolor{red!15}37.29 & \cellcolor{red!15}50.62 & \cellcolor{red!15}64.06 & \cellcolor{red!15}72.01 &\cellcolor{red!15} 33.03 &\cellcolor{red!15} 62.00 &\cellcolor{red!15} 42.55\\ \hline\hline
        \multirow{17}{*}{\em Resnet50}&\multirow{13}{*}{IN-1k$^+$}&\multirow{13}{*}{Yes}&InsDis  & 36.87 & 68.46 & 68.78 & 83.44 & - & 63.39 & 28.98\\
        &  &  &MoCo    & 35.55 & 68.83 & 69.84 & 82.10 & - & 62.10 & 27.99\\
        &  &  &PCL.    & 21.61 & 62.87 & 75.34 & 64.73 & - & 48.02 & 12.93\\
        &  &  &PIRL    & 37.08 & 68.99 & 71.36 & 83.60 & - & 64.65 & 28.72\\
        &  &  &PCLv2   & 37.03 & 70.59 & 82.79 & 85.34 & - & 64.88 & 30.51\\
        &  &  &SimCLR  & 44.90 & 74.20 & 83.33 & 90.87 & - & 67.47 & 43.73\\
        &  &  &MoCov2  & 41.79 & 73.88 & 83.30 & 90.07 & - & 68.95 & 39.31\\
        &  &  &SimCLRv2& 46.38 & 76.38 & 84.72 & 92.90 & - & 73.08 & 50.37\\
        &  &  &SeLav2  & 37.29 & 74.15 & 83.22 & 90.22 & - & 71.08 & 36.86\\
        &  &  &InfoMin & 38.58 & 74.73 & 86.24 & 87.18 & - & 69.53 & 41.01\\
        &  &  &BYOL    & 53.87 & 76.91 & 89.10 & 94.50 & - & 73.01 & 56.40\\
        &  &  &DeepClusterv2& 54.49 & 78.62 & 89.36 & 94.72 & - & 77.94 & 58.60\\
        &  &  &Swav    & 54.04 & 77.02 & 87.60 & 94.62 & - & 76.62 & 54.06\\ \cline{2-11}
        &None&No&DIET& \cellcolor{red!15}44.81 & \cellcolor{red!15}51.75 & \cellcolor{red!15}67.08 & \cellcolor{red!15}73.32 & \cellcolor{red!15}41.03 & \cellcolor{red!15}71.58 & \cellcolor{red!15}55.82\\ \cline{2-11}
        &Tiny&Yes&DIET& \cellcolor{red!15}32.77 & \cellcolor{red!15}49.34 & \cellcolor{red!15}54.56 & \cellcolor{red!15}67.89 & \cellcolor{red!15}24.98 & \cellcolor{red!15}80.43 & \cellcolor{red!15}34.45\\ \cline{2-11}
        &IN100&Yes&DIET& \cellcolor{red!15}36.20 & \cellcolor{red!15}53.85 & \cellcolor{red!15}59.77 &\cellcolor{red!15} 77.22 & \cellcolor{red!15}30.31 & \cellcolor{red!15}80.92 & \cellcolor{red!15}39.96\\\cline{2-11}
        &IN-1k&Yes&DIET& \cellcolor{red!15}35.58 & \cellcolor{red!15}53.48 & \cellcolor{red!15}60.81 &\cellcolor{red!15} 77.16 & \cellcolor{red!15}32.11 & \cellcolor{red!15}81.07 & \cellcolor{red!15}41.12\\ \hline
        \multirow{1}{*}{\em SwinTiny}&None&No&DIET& \cellcolor{red!15}33.15 & \cellcolor{red!15}51.88 & \cellcolor{red!15}58.06 & \cellcolor{red!15}70.78 & \cellcolor{red!15}32.11 & \cellcolor{red!15}68.86 &\cellcolor{red!15} 47.12\\\hline
        \multirow{1}{*}{\em ConvnextSmall}&None&No&DIET& \cellcolor{red!15}43.13 & \cellcolor{red!15}49.52 & \cellcolor{red!15}61.72 & \cellcolor{red!15}67.72 & \cellcolor{red!15}31.44 & \cellcolor{red!15}69.84 &\cellcolor{red!15} 40.63\\
    \end{tabular}
    \label{tab:transfer}
    \vspace{-0.5cm}
\end{table*}

{\bf Small datasets with and without pre-training:}~We conclude the first part of our empirical validation by considering datasets that are commonly handled by SSL through transfer learning: Aircraft \cite{maji13fine-grained}, DTD \cite{cimpoi14describing}, Pets \cite{parkhi2012cats}, Flowers \cite{nilsback2008automated}, CUB200 \cite{WahCUB_200_2011}, Food101 \cite{bossard14}, Cars \cite{KrauseStarkDengFei-Fei_3DRR2013}, where the numbers of training samples and classes are given in \cref{tab:transfer}. The goal is to apply DIET directly on those dataset, without any pre-training, a feast that as far as we are aware, SSL was not able to perform successfully. We thus report those performances in \cref{tab:transfer}. We see that DIET competes with SSL pre-trained models in most of the cases, and arrives not far behind for very small dataset which are DTD, Pets, Flower which contain 1880, 2940 and 1020 training samples respectively. Even then, we see that IN100 pre-training and Resnet18 is near DIET's performances, and that a stronger gap only appears using SSL pre-training on imagenet-1k with a Resnet50.

We additionally propose in \cref{fig:ablation_short} the direct comparison of DIET with supervised learning on a variety of models and dataset but with controlled training size. We clearly observe that for small dataset {\em i.e.} for which we only use a small part of the original training set, DIET is matching supervised performances, which can be considered as ideal since in this setting the data dataset and task is used for evaluation (on the full evaluation dataset).


\subsection{DIET's Dependency on Data-Augmentation, Training Time and Optimizer}
\label{sec:ablation}

We hope in this section to better inform practitioners on the role of Data-Augmentations (DA), training time, and label smoothing into DIET's performances; as well as sensitivity to mini-batch size, which is crucial for single-GPU training.

\begin{table}[t!]
    \centering
    \setlength{\tabcolsep}{0.28em}
    \caption{{\bf TinyImagenet/Resnet18/single-GPU batch-size sensitivity:} top1 test accuracy after a fixed number of epochs (3k) and a single learning rate which is adjusted for each batch-size using $lr=0.001\frac{\texttt{bs}}{256}$; therefore performing per batch-size cross-validation will improve the reported numbers. We observe that DIET is able to perform well across the spectrum of batch sizes and will a slight drop in performance for very small batch size (8, 16) which is by batch-normalization as \cite{ioffe2017batch} identified $32$ to be the minimal working batch-size. Due to large batch sizes being faster (see \cref{fig:BS_CV} for training times), this should be the preferred solution unless hardware constraint is present.}
    \label{tab:BS_CV}
    \begin{tabular}{l|rrrrrrrrr}
     bs&  8   & 16   & 32   & 64   & 128   & 256   & 512   & 1024   & 2048   \\ \hline
     acc.&32.9 & 37.9 & 42.7 & 43.4 &  43.3 &  43.7 &  43.7 &   42.6 &   41.2 \\
    \end{tabular}
\end{table}

{\bf Batch-size does not impact performances.}~One important question when it comes to training a method with low resources is the ability to employ (very) small mini-batch sizes. This is in fact one reason hindering at the deployment of SSL methods which require quite large mini-batch sizes to work (256 is a strict minimum in most cases). We thus propose in \cref{tab:BS_CV} a small sensitivity analysis where we vary the mini-batch size from $8$ to $2048$ --without any tuning of the hyper-parameters-- we use the standard learning rate scaling used in supervised learning: $lr=0.001\frac{\texttt{bs}}{256}$. We observe small fluctuations of performances (due to a sub-optimal learning rate) but no significant drop in performance, even for mini-batch size of $32$. When going to $16$ and $8$, we observe slightly lower performances which we believe to emerge from batch-normalization \cite{ioffe2015batch} which is known to behave erratically below a mini-batch size of $32$ \cite{ioffe2017batch}.

{\bf Data-Augmentations sensitivity is similar to SSL.}~We observed in the previous \cref{sec:SOTA} that when using DA, the proposed DIET was able to almost match high engineered state-of-the-art methods, which should reassure the reader on the usefulness of the method. Yet, knowing which DA to employ is not trivial e.g. many data modalities have no know DA. One natural question is thus around the sensitivity of DIET's performance to the employed DA. To that end, we propose  three DA regimen, one only consistent of random crop and horizontal flips ({\bf strength:1}), which could be consider minimal in computer vision, one which adds color jittering and random grayscale ({\bf strength:2}), and one last which further adds Gaussian blur and random erasing \cite{zhong2020random} ({\bf strength:3}); the exact parameters for those transformations are given in \cref{algo:DA}. We observe on TinyImagenet and with a Resnet34 the following performances 32.93$\pm$ 0.6, 45.60$\pm$ 0.2, and  45.75$\pm$ 0.1 respectively over 5 independent runs, details and additional architectures provided in \cref{fig:DA,tab:DA} in the Appendix. We thus observe that while DIET greatly benefit from richer DA (strength:1 $\mapsto$ 2), it however does not require heavier transformation such as random erasing.

{\bf Convergence is slower than SSL but label smoothing helps.} One important difference in training behavior between supervised learning and SSL is in the number of epochs required to see the quality of the representation plateau. Due to the supervised loss used in DIET, one might wonder how is the training behavior in our case. We observe that the convergence speed of DIET sometimes on-par but often slower than that of SSL in term of number of epochs required to reach a plateau --at least without using label smoothing. In fact, we surprisingly observe that by enabling large values of label smoothing, {\em e.g.} $0.8$, it was possible to obtain faster convergence. We provide a sensitivity analysis in \cref{fig:epochs,tab:epochs} in the Appendix. We believe that convergence speed could be improved by designing improved update mechanism for DIET's linear classifier. In fact, one should recall that within a single epoch, only one of each datum/class is observed, making the convergence speed of the classifier's {\color{red}$\mW$} matrix the main limitation; we hope to explore improved training strategies in the future as discussed in \cref{sec:conclusion}.

\subsection{Pushing the DIET to Large Models and Datasets}
\label{sec:scaling}

Given DIET's formulation of considering each datum as its own class, it is natural to ask ourselves how scalable is such a method. Although we saw that on small and medium scale dataset, DIET's was able to come on-par with most current SSL methods, it is not clear at all it remains true for larger datasets. In this section we briefly describe what can be done to employ DIET on datasets such as Imagenet and INaturalist.

The first dataset we consider is INaturalist which contains slightly more than $500K$ training samples for its mini version (the one commonly employed, see {\em e.g. }\cite{zbontar2021barlow}). It contains almost $10K$ actual classes and most SSL methods focus on transfer learning {\em e.g.} transferring with a Resnet50 from Imagenet-1k lead to SimCLR's 37.2$\%$, MoCoV2's 38.6, BYOL's 47.6 and BarlowTwins'46.5. However training on INaturalist directly produces lower performances reaching only 29.1 with MSN and a ViT. Using DIET is possible out-of-the-box with Resnet18 and ViT variants as their embedding is of dimension 512 and 762 respectively making $\mW$ fit in memory. We obtain 22.81 with a convnext small, and 21.6 with a ViT.

The second dataset we consider is the full Imagenet-1k dataset which contains more than 1 million training samples and 1000 actual classes. In this case, it is not possible to directly hold $\mW$ in-memory. We however tried a simple strategy which simply consists of sub-sampling the training set to a more reasonable size. This means that although we are putting aside many training images, we enable single GPU Imagenet training with DIET. With a training size of $400K$, we able to reach 44.05 with a convnext small, 43.78 with a SwinTiny, and 44.89 with a ViT/B/16. A standard SSL pipeline has performances ranging between $64\%$ and $72\%$. From those experiments, it is clear that DIET's main limitation comes from very large training set sizes. Although the above simple strategy offers a workable solution, it is clearly not sufficient to match with existing unsupervised learning method and thus should require further consideration. As highlighted in \cref{sec:conclusion} below, this is one key avenue for future work.

\input{content/conclusion}


\bibliography{bibliography}
\bibliographystyle{icml2023}


\appendix
\onecolumn
\input{content/proof}

\end{document}

%% file: content/abstract.tex
Costly, noisy, and over-specialized, labels are to be set aside in favor of unsupervised learning if we hope to learn cheap, reliable, and transferable models. To that end, spectral embedding, self-supervised learning, or generative modeling have offered competitive solutions. Those methods however come with numerous challenges \textit{e.g.} estimating geodesic distances, specifying projector architectures and anti-collapse losses, or specifying decoder architectures and reconstruction losses. In contrast, we introduce a simple explainable alternative ---coined \textbf{DIET}--- to learn representations from unlabeled data, free of those challenges. \textbf{DIET} is blatantly simple: take one's favorite classification setup and use the \textbf{D}atum \textbf{I}nd\textbf{E}x as its \textbf{T}arget class, \textit{i.e. each sample is its own class}, no further changes needed. \textbf{DIET} works without a decoder/projector network, is not based on positive pairs nor reconstruction, introduces no hyper-parameters, and works out-of-the-box across datasets and architectures. Despite \textbf{DIET}'s simplicity, the learned representations are of high-quality and often on-par with the state-of-the-art \textit{e.g.} using a linear classifier on top of DIET's learned representation reaches $71.4\%$ on CIFAR100 with a Resnet101, $52.5\%$ on TinyImagenet with a Resnext50.

%% file: content/introduction.tex
\section{Introduction}
\label{sec:introduction}

{\em Unsupervised learning} of a model $f_{\vtheta}$, governed by some parameter $\vtheta$, has always been and still is one of the most challenging and rewarding task in deep learning \cite{bengio2012deep}. In fact, {\em supervised learning} which learns to produce predictions from known input-output pairs can be considered solved in contrast to unsupervised learning which aims to produce descriptive and intelligible representations from inputs only \cite{hastie2009overview,goodfellow2016deep}. 

{\em Self-Supervised Learning} (SSL) \citep{chen2020simple,misra2020self} has recently demonstrated that one could learn without labels  highly non-trivial Deep Networks (DNs) whose representations are as descriptive as supervised ones. In particular, SSL differs from {\em reconstruction-based} methods such as (denoising, variational) Autoencoders \citep{vincent2008extracting,vincent2010stacked,kingma2013auto} and their cross-variants by removing the need for a {\em decoder} DN and an input-space reconstruction loss, both being difficult to design \citep{wang2004image,grimes2005bilinear,larsen2016autoencoding,cosentino2022spatial}. SSL's recent developments have also led to outperforming {\em reconstruction-free} methods {\em e.g.} Noise Contrastive Estimation and its variants \citep{hyvarinen2005estimation,hinton2002training,song2019generative,rhodes2020telescoping}. 
Nonetheless, SSL which is the current state-of-the-art unsupervised learning solution, comes with many moving pieces {\em e.g.} (i) artificially constructed {\em positive pairs}, commonly obtained by applying two a priori known and tailored Data-Augmentations (DAs) to each datum, (ii) a carefully designed {\em projector} DN $g_{\vgamma}$ to perform SSL training with the composition $g_{\vgamma} \circ f_{\vtheta}$ and throwing away the projector $g_{\vgamma}$ afterwards \cite{chen2020simple}, or (iii) advanced anti-collapse techniques involving moving average teacher models \cite{grill2020bootstrap,caron2021emerging}, representation normalization \cite{chen2021exploring,zbontar2021barlow}, or Entropy estimation \cite{chen2020simple,li2021self}. An incorrect pick of any of those moving pieces results in a drastic drop in performances \cite{cosentino2022toward,bordes2022guillotine}. This design sensitivity poses a real challenge as SSL is computationally demanding therefore preventing cross-validation to be employed at every change in the DN's architecture and/or dataset. Even more limiting, SSL's cross-validation relies on assessing the quality of the produced DN through the dataset's labels and test accuracy {\em e.g.} from a (supervised) linear probe. This supervised quality assessment is required simply because the actual values of current SSL losses fail to convey any qualitative information about the representation being learned \cite{ghosh2022investigating,garrido2022rankme}.

Instead of further refining of existing SSL methods, and thus inheriting most of their limitations, we propose a stripped down unsupervised learning solution, free of all those challenges ---coined \textbf{DIET}. Unsupervised learning on a \textbf{DIET} consists in exploiting the most stable and understood setting: {\em supervised learning} but removing the need for labels by instead {\em using the \textbf{D}atum \textbf{I}nd\textbf{E}x as its \textbf{T}arget label}. Three striking observations will emerge. First, this simple \textbf{DIET} learns high-quality representations that are surprisingly on-par with much more sophisticated state-of-the-art methods. Second, the \textbf{DIET} performs equally well regardless of the dataset and architecture being considered. This is in contrast with most existing methods that are often architecture specific. Lastly, and perhaps of most importance, the \textbf{DIET} can be employed with low-resources {\em e.g.} most of our experiments employ a single GPU, and DIET's training loss is informative of downstream performances. Again, since \textbf{DIET} is a supervised learning method with the datum index as target, its training is as stable as the current state-of-the-art supervised methods and any progress made within the supervised learning realm is directly transferable to the proposed \textbf{DIET}. We hope that our method provides a novel go-to solution for many practitioners interested in learning high-quality representations, in a stable and minimalist way.
\\
We summarize our contributions below:
\begin{enumerate}[itemsep=-4pt,topsep=-4pt]
    \item We introduce a \textbf{DIET} for unsupervised learning in \cref{sec:DIET} (summarized in \cref{fig:DIET}), a competitive and minimalist strategy bloating a few key benefits\dots
    \item {\bf Stable and Out-of-the-box:} we validate the DIET on 16 official DN architectures including ConvNexts or ViTs, and on 10 datasets; the same setup (\cref{fig:hparams}) is successful across all cases (\cref{tab:C100,tab:tiny,tab:transfer})
    \item {\bf No hyper-parameter and single-GPU:} moving away from decoders/positive pairs/projectors/\dots allows us to propose a DIET introducing no hyper-parameter and requiring only a few lines of code (\cref{algo:DIET}) and we perform sensitivity analysis on batch size, data-augmentation, training time in \cref{sec:ablation}
    \item {\bf Informative training loss:} the DIET's training loss strongly correlates with the test accuracy across architectures and datasets (\cref{fig:accus}) enabling informed cross-validation to unsupervised learning
\end{enumerate}
Code provided in \cref{algo:DIET}.

%% file: content/background.tex
\begin{figure*}[t!]
\centering    
\begin{tikzpicture}
\node[inner sep=0pt] (a) at (0,0) {\includegraphics[width=2cm,height=2cm]{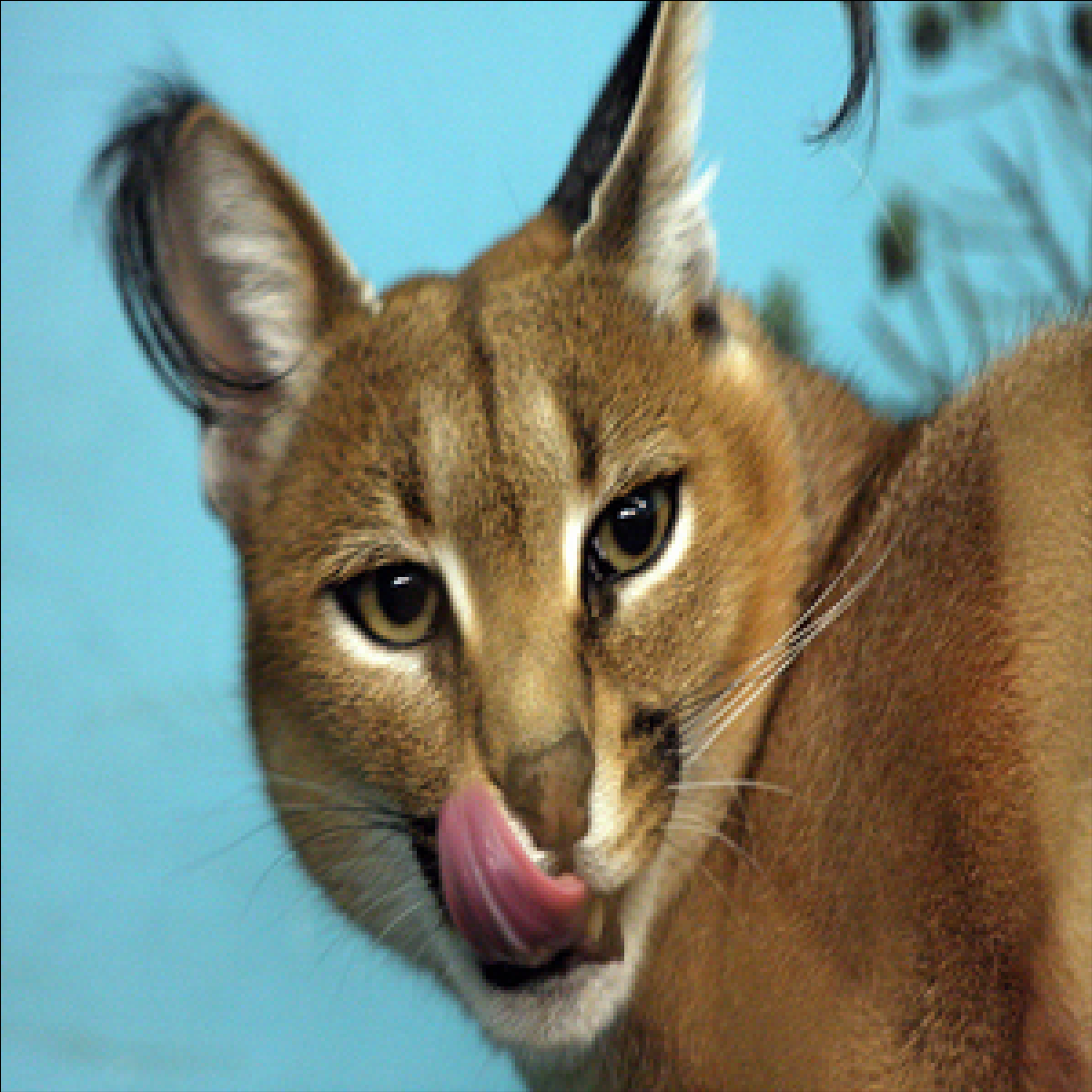}};
\node[inner sep=0pt,above=.1cm of a]  {$\texttt{datum}_{\texttt{1}}$};

\node[inner sep=0pt,right=.2cm of a] (b) {\includegraphics[width=2cm,height=2cm]{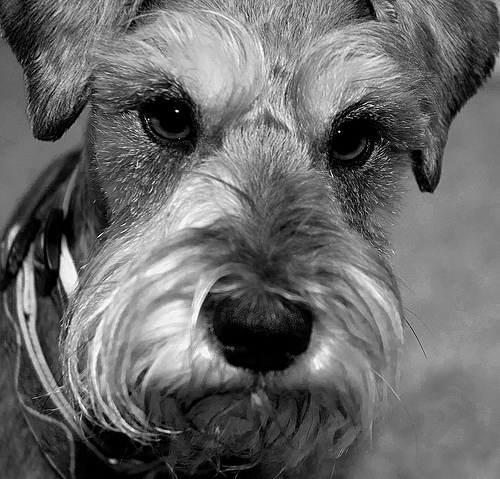}};
\node[inner sep=0pt,above=.1cm of b]  {$\texttt{datum}_{\texttt{2}}$};
\node[inner sep=0pt,above right=.4cm and -2cm of b]  {\textbf{Training dataset}};

\node[inner sep=0pt,right=.2cm of b] (c) {\parbox{0.7cm}{\centering\dots}};

\node[inner sep=0pt,right=.2cm of c] (d) {\includegraphics[width=2cm,height=2cm]{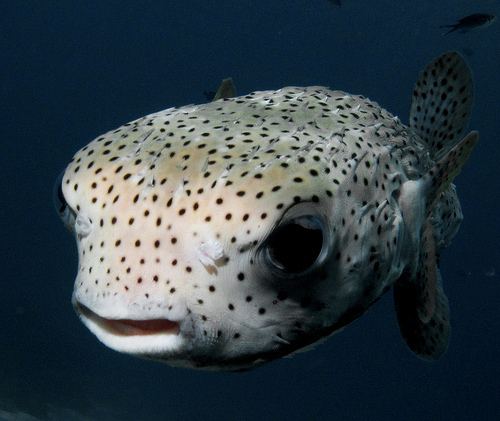}};
\node[inner sep=0pt,above=.1cm of d]  {$\texttt{datum}_{\texttt{N}}$};
    
\draw[black,thick,rounded corners] ($(a.north west)+(-0.1,0.8)$)  rectangle ($(d.south east)+(0.1,-0.1)$);

\node[inner sep=0pt,below right=-.8cm and 0.5cm of d] (node2) {\parbox{2cm}{\centering$\mathcal{T}(\texttt{datum}_{\texttt{n}})$\\{\scriptsize (with DA $\mathcal{T}$)}}};
\node[inner sep=0pt,above=2.2cm of node2] {sample a datum index (\texttt{n})};

\definecolor{red}{RGB}{207,78,83}
\definecolor{blue}{RGB}{87,148,160}
\tikzstyle{every pin edge}=[<-,shorten <=1pt]
    \tikzstyle{neuron}=[circle,fill=black!25,minimum size=17pt,inner sep=0pt]
    \tikzstyle{co}=[black!55]
    \tikzstyle{input neuron}=[neuron, fill=black!60];
    \tikzstyle{output neuron}=[neuron, fill=red];
    \tikzstyle{hidden neuron}=[neuron, fill=blue];

    \node[input neuron,right=3.2cm of d] (I-2){};
    \node[input neuron,above=0.2cm of I-2] (I-1){};
    \node[input neuron,below=0.2cm of I-2] (I-3){};

    \node[hidden neuron,right=0.4cm of I-2] (H-3) {};
    \node[hidden neuron,above=0.3cm of H-3] (H-4) {};
    \node[hidden neuron,above=0.3cm of H-4] (H-5) {};
    \node[hidden neuron,below=0.3cm of H-3] (H-2) {};
    \node[hidden neuron,below=0.3cm of H-2] (H-1) {};
    
    \node[hidden neuron,below right=0.1cm and 0.6cm of H-4] (h-2) {};
    \node[hidden neuron,below=0.3cm of h-2] (h-1) {};

    \node[output neuron, above right=2.5cm and 1.cm of h-1] (O1) {};

    \foreach \source in {1,...,3}
        \foreach \dest in {1,...,5}
            \path[blue!70] (I-\source) edge (H-\dest);
    \foreach \dest in {1,...,5}{
        \path[blue!70] (H-\dest) edge (h-1);
        \path[blue!70] (H-\dest) edge (h-2);
        }
            
    \foreach \source in {1,...,2}
        \path[red!70] (h-\source) edge (O1);
    \foreach \dest/\new in {1/2,2/3,3/4,4/5,5/6,6/7}{
        \node[output neuron, below=0.2cm of O\dest] (O\new) {};
        \foreach \source in {1,...,2}
            \path[red!70] (h-\source) edge (O\new);
    }
    
\draw [pen colour={red},thick,decorate,decoration={calligraphic brace,amplitude=2mm, raise=1.5mm}] ($(O1.north east)+(-1.4,0.)$) -- (O1.north east);
\node[red,above right=0.3cm and -2.1cm of O1] {\parbox{2.5cm}{\centering\texttt{N}-output linear\\layer (logits)}};

\draw [pen colour={blue},thick,decorate,decoration={calligraphic brace,amplitude=2mm, raise=1.5mm,mirror}] ($(H-1.south west)-(0.7,0)$) -- ($(H-1.south west)+(1.6,0)$);
\node[blue,below right=0.3cm and -1.1cm of H-1] {Deep Network};
    
\node[inner sep=0pt,right=1cm of O4] (TO){\color{red}$\vy_n$}; 
\path[->,red!70,line width=0.8mm] ($(O4.east)+(0.2,0)$) edge ($(TO.west)+(-0.1,0)$);
\node[inner sep=0pt,below=.5cm of TO] (TO2) {\texttt{X-Ent}$(\texttt{n},$\color{red}$\vy_\texttt{n}$\color{black}$)$};
\path[->,black!70,very thick] (TO) edge (TO2);
\path[->,black!70,line width=0.8mm] ($(d.east)+(0.3,0.)$) edge ($(I-2)+(-0.5,0)$);
\path[->,black!70,very thick] (TO) edge (TO2);
\draw[black!70,line width=0.8mm,rounded corners=0.3cm] ($(d.east)+(0.8,1.9)$) -- node[midway] {} ($(d.east)+(0.8,0.0)$) -- node[midway] {}($(d.east)+(1.4,0.0)$);

\node[above right=2.5cm and -2.1cm of a] {\parbox{9.2cm}{\Large Datum IndEx as Target (\textbf{DIET})}};

\node[above left=1.1cm and -9cm of a] (top1) {
\parbox{9cm}{\small\em
\begin{itemize}[itemsep=-5pt,topsep=0pt]
    \item no siamese/teacher-student/projector DN
    \item no representation collapse
    \item informative training loss
    \item out-of-the-box across architectures/datasets
\end{itemize}
}};

\end{tikzpicture}\vspace{-0.6cm}
    \caption{\textbf{DIET} uses the datum index (\texttt{n}) as the class-target --effectively turning unsupervised learning into a supervised learning problem. In our case, we employ the cross-entropy loss (\texttt{X-Ent}), no extra care needed to handle different dataset or architectures. As opposed to current SOTA, we do not rely on a projector nor positive views \textit{i.e} no change need to be done to any existing supervised pipeline to obtain the DIET. As highlighted in \cref{fig:accus}, DIET's training loss is even informative of downstream test performances, and as ablated in \cref{sec:ablation} there is no degradation of performance with longer training, even for very small datasets (\cref{tab:transfer}).}
    \label{fig:DIET}
\end{figure*}

\section{Why Unsupervised Learning Needs a DIET}
\label{sec:background}

The current state of unsupervised learning consists in complicated methods combining numerous moving pieces that need re-tweaking for each DN architecture and dataset. As a result reproducibility, scalability, and explainability are hindered.

{\bf Spectral embedding is untractable.}~Spectral embedding takes many forms but can be summarized into estimating geodesic distances \cite{meng2008improving,thomas2013geodesic} between all or some pairs of training samples to then learn a non-parametric \cite{roweis2000nonlinear,belkin2001laplacian,balasubramanian2002isomap,brand2003unifying}, or parametric \cite{bengio2003out,pfau2018spectral} mapping that produces embeddings whose pairwise $\ell_2$ distance matches the estimated geodesic ones. As such, spectral embedding heavily relies on the estimation of the geodesic distances which is a challenging problems \cite{lantuejoul1981use,lantuejoul1984geodesic,revaud2015epicflow}, especially for images and videos \cite{donoho2005image,wakin2005high}. This limitation fueled the development of alternative methods \textit{e.g.} Self-Supervised Learning (SSL), that often employ similar losses than spectral embedding \cite{haochen2021provable,balestriero2022contrastive,cabannes2022minimal} but manage to move away from geodesic distance estimation through the explicit generation of positive pairs \textit{i.e.} samples with known geodesic distances.

{\bf Self-Supervised Learning is unintelligible.}~Despite flamboyant performance reporting and well motivated first principles, SSL --as of today-- falls back to combining numerous hacks driven by supervised performances. In fact, SSL has evolved to a point where novel methods are architecture specific. A few challenges that limit SSL to be widely adopted are (i) loss values which are uninformative of the DN's quality \citep{reed2021selfaugment,garrido2022rankme}, partly explained by the fact that SSL composes the DN of interest $f_{\theta}$ with a projector DN $ g_{\gamma}$ appended to it during training and thrown away afterward, (ii) too many per-loss and per-projector hyper-parameters whose impact on the DN's performances are hard to control or predict \citep{grill2020bootstrap,tian2021understanding,he2022exploring}, and which are even widely inconsistent across datasets and architectures \citet{zhai2019visual,cosentino2022geometry}, (iii) lack of theoretical guarantees as all existing studies have derived optimality conditions at the projector's output \citep{wang2020understanding,tian2020makes,jing2021understanding,huang2021towards,haochen2021provable,dubois2022improving,zhang2021understanding,wang2021understanding} which is not the output of interest since the projector is thrown away after SSL training, and it is known that the DN's output and the projector's output greatly differ, see e.g. Tab.~3, Tab.~1, Tab.~1 and Fig.~1 of \citep{chen2020simple,chen2020improved,cosentino2022toward,bordes2022guillotine} respectively.
From a more practical standpoint, SSL requires to generate positive pairs making it much more costly and resource hungry than standard supervised learning.

{\bf Reconstruction-based learning is unstable.}~
Reconstruction without careful tuning of the loss has been known to be sub-optimal for long \citep{bishop1994mixture,graves2013generating} and new studies keep reminding us of that \citep{lecun2022path}. The argument is simple, suppose one aims to minimize a reconstruction metric $R$ for some input $\vx$
\begin{align}
    R(d_{\gamma}(e_{\eta}(\vx)),\vx),\label{eq:r}
\end{align}
where $e_{\eta}$ and $d_{\gamma}$ are parametrized learnable encoder and decoder networks respectively; $e_{\eta}(\vx)$ is the representation of interest to be used after training. In practice, as soon as some noise $\epsilon$ is present in the data, \textit{i.e.} we observe $\vx + \epsilon$ and not $\vx$, that noise $\epsilon$ must be encoded by $e_{\eta}$ to minimize the loss from \cref{eq:r} unless one carefully designs $R$ so that 
$R(\vx+\epsilon,\vx) = 0$. However, designing such a {\em noise invariant} $R$ has been attempted for decades \citep{park1995distance,simoncelli1996rotation,fienup1997invariant,grimes2005bilinear,wang2005translation} and remains a challenging open problem. Hence, many solutions rely on learning $R$ e.g. in VAE-GANs \citep{larsen2016autoencoding} bringing even further instabilities and training challenges. Other alternatives carefully tweak $R$ per dataset and architectures e.g. to only compute the reconstruction loss on parts of the data as with BERT \citet{devlin2018bert} or MAEs \citet{he2022masked}. Lastly, the quality of the encoder's representation depends on its architecture but also on the decoder's one \citep{yang2017improved,xu2021improving} making cross-validation more costly and unstable \citep{antun2020instabilities}.

In numerous scenarios one finds themselves in a position where none of the existing state-of-the-art's limitations can be overcome --motivating the development of our {\bf DIET}-- an overly simple yet highly effective unsupervised learning strategy that inherits none of the aforementioned challenges.

%% file: content/conclusion.tex
\section{Conclusions and Future Work}
\label{sec:conclusion}

We presented a simple unsupervised learning method coined DIET, for Datum IndEx as Target, which simply casts the task of descriptive representation learning with Deep Networks (DNs) into a supervised problem of instance discrimination. Despite its simplicity, DIET is able to learn competitive representations that are often on-par with current state-of-the-art methods {\em e.g.} Self-Supervised Learning (SSL). We believe that DIET provides an out-of-the-box solution for many situations since (i) its training loss functions is informative of the downstream task test accuracy, (ii) it does not introduce any additional hyper-parameters, and training works seamlessly across architectures, and (iii) its implementation is requiring nearly no code refactoring from already built supervised pipelines, which contrast with {\em e.g.} SSL or autoencoders for which complicated data pipelines and additional DN specification. That being said, DIET suffers from one main limitations: the computational and memory complexity grows linearly with the dataset size, opening a few avenues.

To speed up training, a smarter initialization of the N-output classifier could be envisioned, along with a possibly different learning schedule for this large matrix and for the rest of the DN.


%% file: content/proof.tex
\newpage

\begin{center}
    \Huge
    Supplementary Materials
\end{center}

The supplementary materials is providing the proofs of the main's paper formal results. We also provide as much background results and references as possible throughout to ensure that all the derivations are self-contained.
Some of the below derivation do not belong to formal statements but are included to help the curious readers get additional insights into current SSL methods.

\section{Code}
\label{sec:code}









\begin{algorithm}[h]
\begin{lstlisting}[language=Python,escapechar=\%,numbers=none]
model = torchvision.models.__dict__[architecture]()

# CIFAR procedure to adjust to the lower image resolution
if is_cifar and "resnet" in architecture:
    model.conv1 = torch.nn.Conv2d(3, 64, kernel_size=3, stride=1, padding=2, bias=False)
    model.maxpool = torch.nn.Identity()

# for each architecture, remove the classifier and get the output dim. (K)
if "alexnet" in architecture:
    K = model.classifier[6].in_features
    model.classifier[6] = torch.nn.Identity()
elif "convnext" in architecture:
    K = model.classifier[2].in_features
    model.classifier[2] = torch.nn.Identity()
elif "convnext" in architecture:
    K = model.classifier[2].in_features
    model.classifier[2] = torch.nn.Identity()
elif "resnet" in architecture or "resnext" in architecture or "regnet" in architecture:
    K = model.fc.in_features
    model.fc = torch.nn.Identity()
elif "densenet" in architecture:
    K = model.classifier.in_features
    model.classifier = torch.nn.Identity()
elif "mobile" in architecture:
    K = model.classifier[-1].in_features
    model.classifier[-1] = torch.nn.Identity()
elif "vit" in architecture:
    K = model.heads.head.in_features
    model.heads.head = torch.nn.Identity()
elif "swin" in architecture:
    K = model.head.in_features
    model.head = torch.nn.Identity()
\end{lstlisting}
\caption{\small Get the output dimension and remove the linear classifier from a given torchvision model (Pytorch used for illustration).}
\label{algo:architectures}
\end{algorithm}

\begin{figure}[t!]
    \centering
    \includegraphics[width=\linewidth]{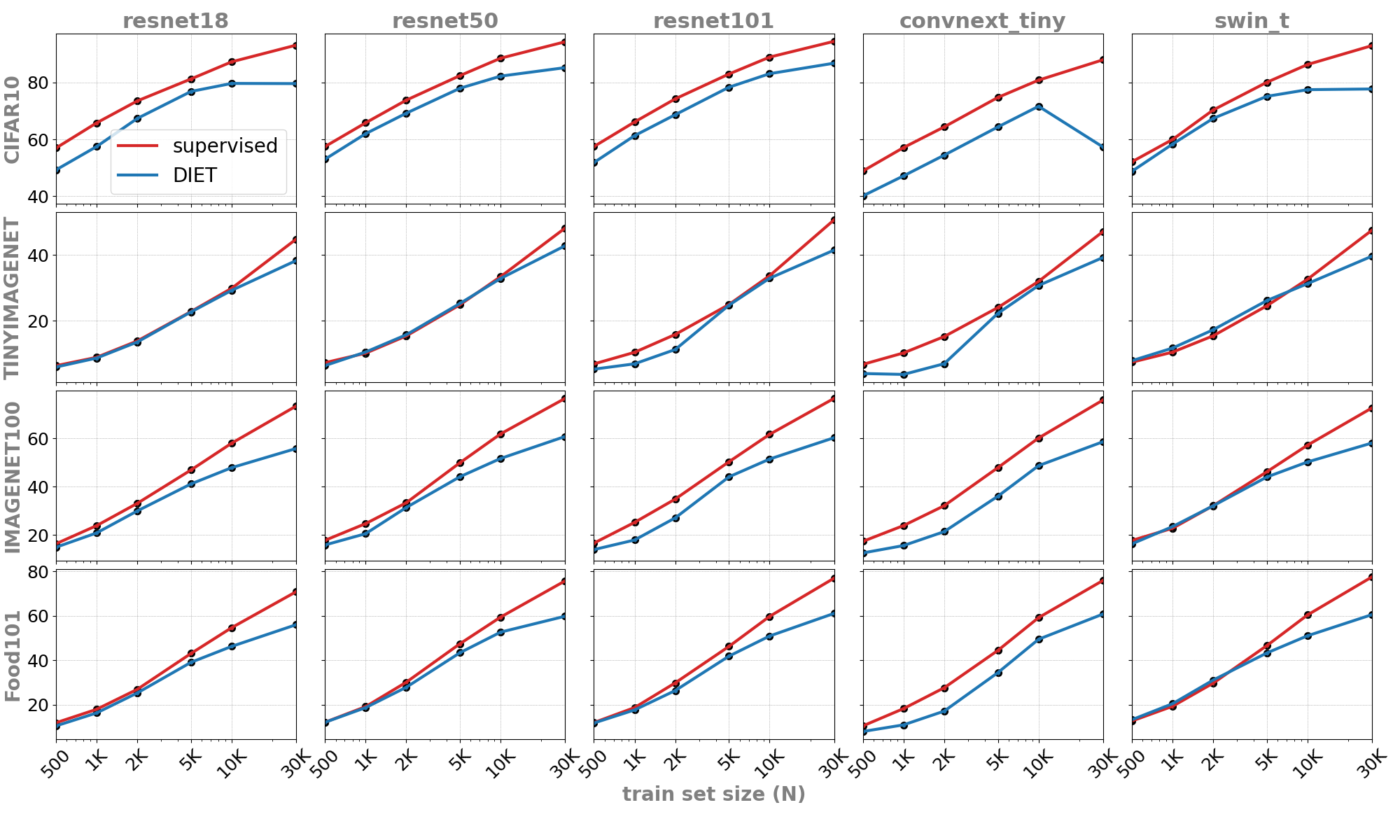}
    \caption{Reprise of \cref{fig:ablation_short} but on additional datasets.}
    \label{fig:ablation}
\end{figure}

\section{Impact of Training Time and Label Smoothing}
\begin{figure}[h]
    \centering
    Resnet18\\
    \includegraphics[width=\linewidth]{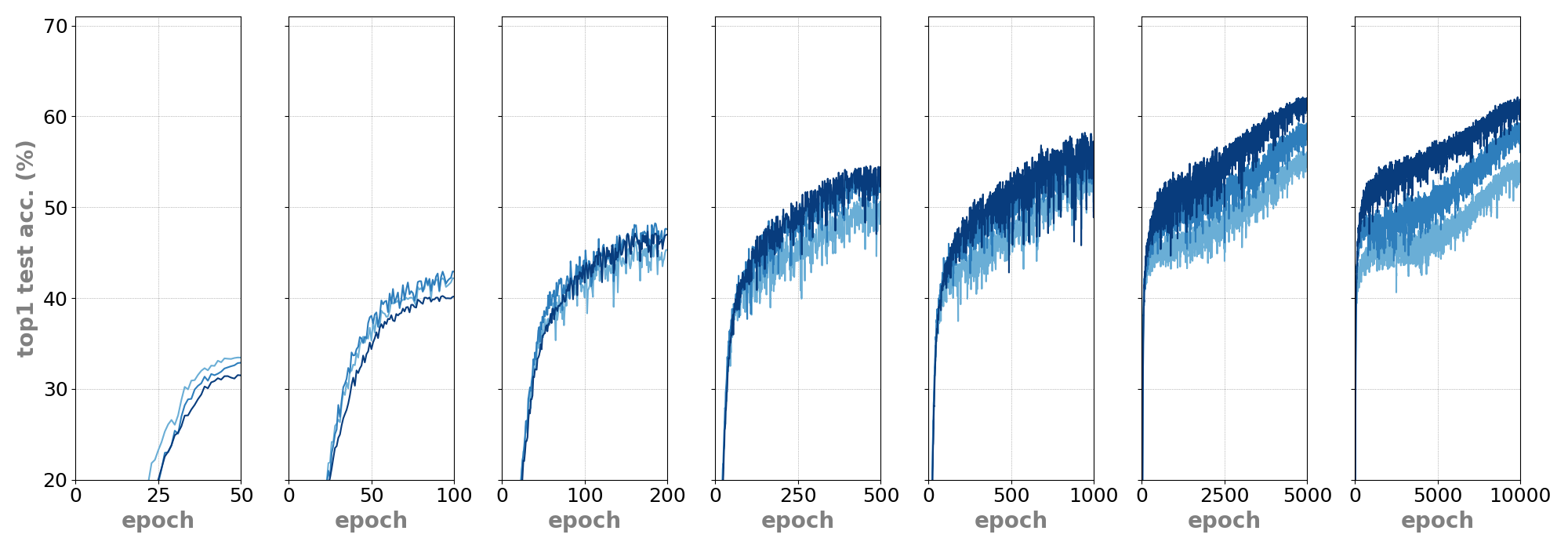}\\
    Resnet50\\
    \includegraphics[width=\linewidth]{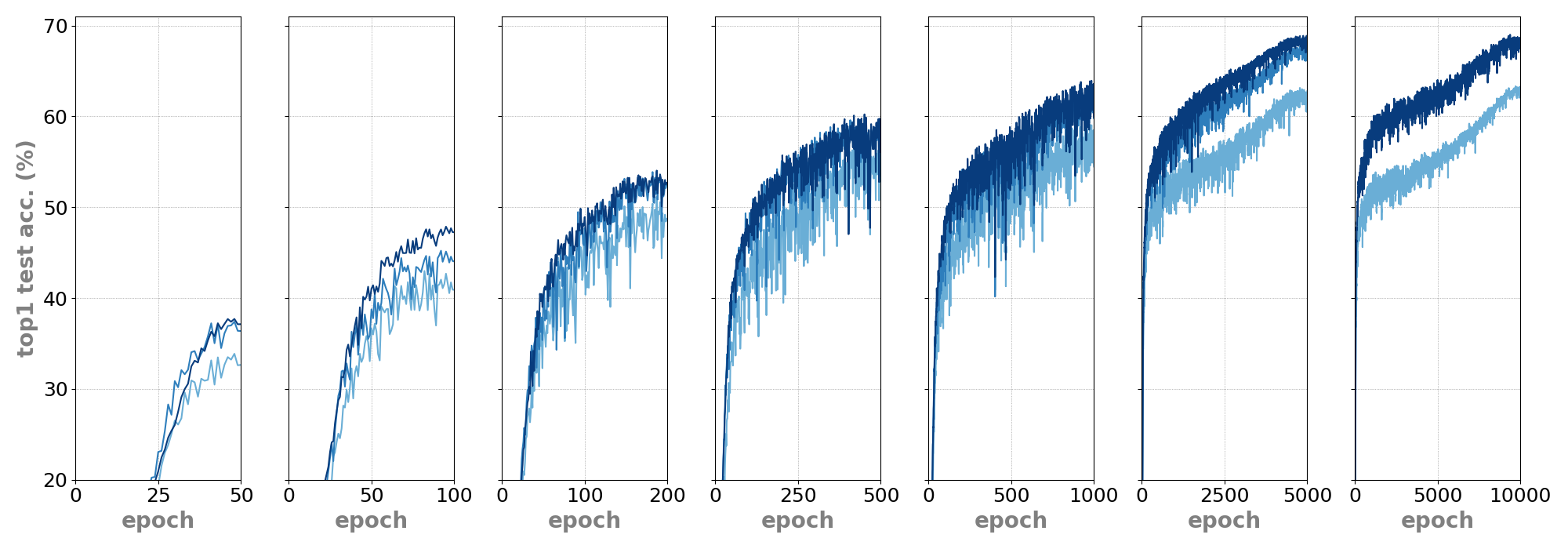}\\
    Resnet101\\
    \includegraphics[width=\linewidth]{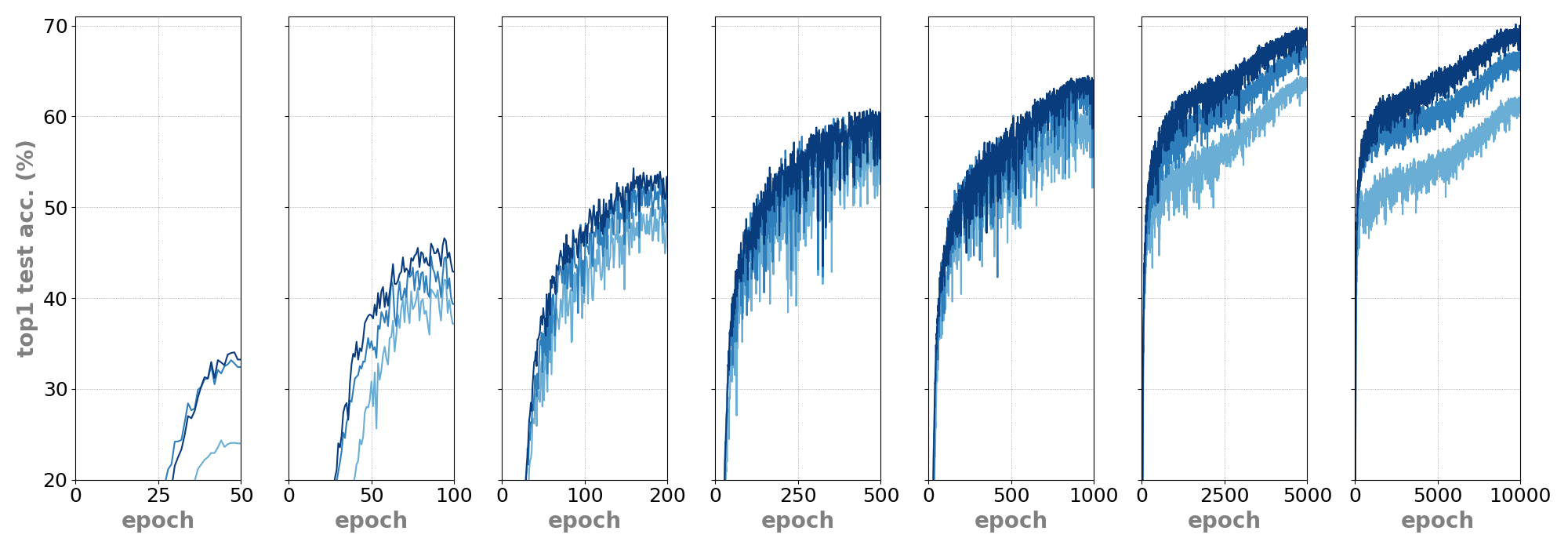}
    \caption{Depiction of the evolution of linear top1 accuracy throughout epochs on CIFAR100 with three Resnet variants and three label smoothing parameters represented by the different {\bf shades of blue} going from light to dark shades with values of $0.1$, $0.4$, and $0.8$ respectively. We clearly observe that higher value of label smoothing speeds up convergence, although in this setting all cases greatly benefit from longer training schedules; final linear probe performances are reported in \cref{tab:epochs}.}
    \label{fig:epochs}
\end{figure}

\begin{table}[h]
    \centering
    \caption{Final top1 test accuracy values on CIFAR100 with the models and training schedules depicted in \cref{fig:epochs}, only the label smoothing value with the best accuracy is reported.}
    \label{tab:epochs}
    \begin{tabular}{lrrrrrrr}
    \hline
     Epochs    & 50    & 100    & 200    & 500    & 1000    & 5000    & 10000    \\
     resnet18  & 33.46 &  42.94 &  48.24 &  54.54 &   58.81 &   62.63 &    63.29 \\
     resnet50  & 37.71 &  47.86 &  54.04 &  60.23 &   64.24 &   69.51 &    69.91 \\
     resnet101 & 34.03 &  46.59 &  54.3  &  60.8  &   64.71 &   70.56 &    71.39 \\
    \hline
    \end{tabular}
\end{table}

\section{Impact of Mini-Batch Size}

\begin{figure}[h]
    \centering
    \begin{minipage}{0.6\linewidth}
    \includegraphics[width=\linewidth]{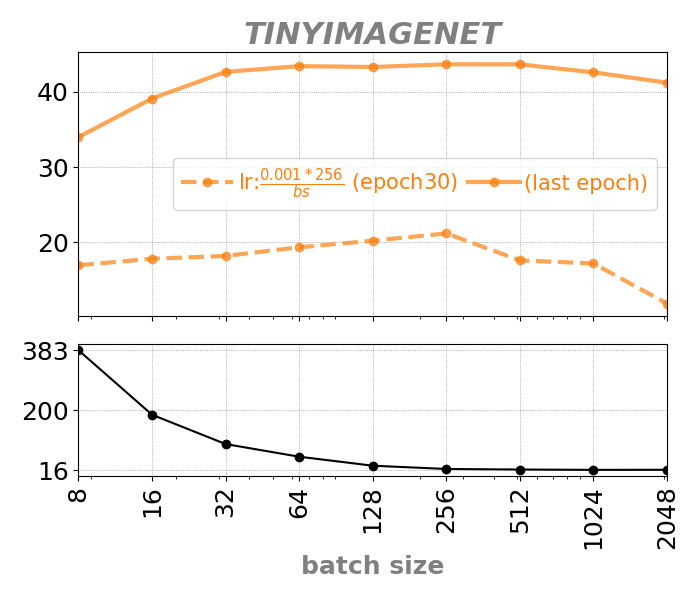}
    \end{minipage}
    \begin{minipage}{0.39\linewidth}
    \caption{TinyImagenet with fixed number of epochs and a single learning rate which is adjusted for each case using the LARS rule therefore per batch-size learning cross-validation can only improve performances, see \cref{tab:BS_CV}, , the per-epoch time includes training, testing, and checkpointing.}
    \label{fig:BS_CV}
    \end{minipage}
\end{figure}

\section{Impact of Data-Augmentation}

\begin{figure}[h]
\begin{minipage}{0.49\linewidth}
    \centering
    \includegraphics[width=\linewidth]{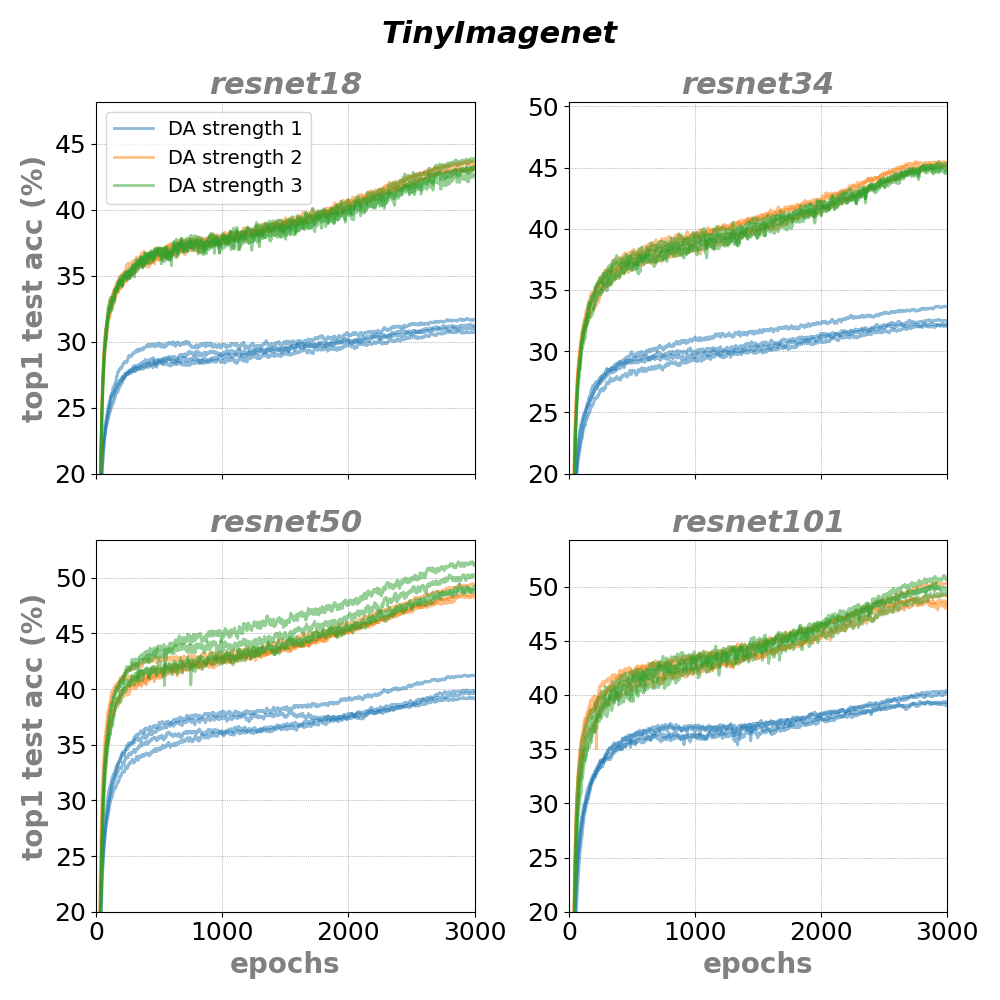}
    \caption{TinyImagenet, see \cref{tab:DA} for table of results, and the specific DAs can be found in \cref{algo:DA}.}
    \label{fig:DA}
\end{minipage}
\begin{minipage}{0.49\linewidth}
\captionsetup{type=table} 
    \centering
    \caption{TinyImagenet top1 test accuracy with a faster training schedule than the \cref{tab:tiny} (2000 epochs) with average and std over $5$ runs with 3 different DA augmentation pipelines and $4$ different architectures, the per-epoch linear probe evaluation performances are reported in \cref{fig:DA}, DAs can be found in \cref{algo:DA}.}
    \label{tab:DA}
    \begin{tabular}{l|lll}
    \hline
     &strength: 1    & strength: 2    & strength: 3    \\
    \hline
    resnet18& 31.48$\pm$ 0.3 & 43.62$\pm$ 0.2 & 43.88$\pm$ 0.3 \\
    resnet34& 32.93$\pm$ 0.6 & 45.60$\pm$ 0.2  & 45.75$\pm$ 0.1 \\
    resnet50& 40.24$\pm$ 0.7 & 48.80$\pm$ 0.6  & 50.81$\pm$ 0.8 \\
    resnet101& 40.07$\pm$ 0.4 & 49.74$\pm$ 0.5 & 50.76$\pm$ 0.5 \\
    \hline
    \end{tabular}
\end{minipage}
\end{figure}

\begin{algorithm}[h]
\begin{lstlisting}[language=Python,escapechar=\%,numbers=none]
transforms = [
    RandomResizedCropRGBImageDecoder((size, size)),
    RandomHorizontalFlip(),
]
if strength > 1:
    transforms.append(
        T.RandomApply(
            torch.nn.ModuleList([T.ColorJitter(0.4, 0.4, 0.4, 0.2)]), p=0.3
        )
    )
    transforms.append(T.RandomGrayscale(0.2))
if strength > 2:
    transforms.append(
        T.RandomApply(
            torch.nn.ModuleList([T.GaussianBlur((3, 3), (1.0, 2.0))]), p=0.2
        )
    )
    transforms.append(T.RandomErasing(0.25))
\end{lstlisting}
\caption{\small Custom dataset to obtain the indices ($n$) in addition to inputs $\vx_n$ and (optionally) the labels $y_n$ to obtain \texttt{train\_loader} used in \cref{algo:DIET} (Pytorch used for illustration).}
\label{algo:DA}
\end{algorithm}

\newpage